\documentclass[letterpaper, 10pt, conference]{IEEEtran}
\usepackage[top=0.75in, bottom=0.78in, left=0.75in, right=0.75in]{geometry}

\IEEEoverridecommandlockouts  

\usepackage{orcidlink}
\usepackage{hyperref}
\usepackage[normalem]{ulem} 
\usepackage{color}
\usepackage{tcolorbox}
\usepackage{listings}
\usepackage{xcolor} 

\lstdefinelanguage{json}{
    basicstyle=\ttfamily\footnotesize,
    numbers=none, 
    numberstyle=\tiny\color{gray},
    showstringspaces=false,
    breaklines=true,
    frameround=ftff,
    keywordstyle=\color{blue}\bfseries,
    stringstyle=\color{teal},
    commentstyle=\color{gray},
    flexiblecolumns=true,
    literate=
     *{0}{{{\color{red}0}}}{1}
      {1}{{{\color{red}1}}}{1}
      {2}{{{\color{red}2}}}{1}
      {3}{{{\color{red}3}}}{1}
      {4}{{{\color{red}4}}}{1}
      {5}{{{\color{red}5}}}{1}
      {6}{{{\color{red}6}}}{1}
      {7}{{{\color{red}7}}}{1}
      {8}{{{\color{red}8}}}{1}
      {9}{{{\color{red}9}}}{1}
      {:}{{{\color{black}:}}}{1}
      {,}{{{\color{black},}}}{1}
      {\{}{\color{black}\textbraceleft}{1}
      {\}}{\color{black}\textbraceright}{1}
      {[}{\color{black}\textlbrackhook}{1}
      {]}{\color{black}\textrbrackhook}{1},
}

\usepackage{soul}
\sethlcolor{lightgray!50}
\usepackage{multirow}
\usepackage{booktabs}
\usepackage{float}

\usepackage{hyperref}
\usepackage{soul} 

\definecolor{lightblue}{rgb}{0.93,0.96,1}
\definecolor{hellgrauHintergrundZwei}{rgb}{0.95, 0.95, 0.95}
\definecolor{dunkelblaugruen60heller}{rgb}{0.6, 0.78, 0.9}

\usepackage{amsmath,amssymb,amsfonts}
\usepackage{algorithmic}
\usepackage{graphicx}
\usepackage{textcomp}
\usepackage{xcolor}
\def\BibTeX{{\rm B\kern-.05em{\sc i\kern-.025em b}\kern-.08em
    T\kern-.1667em\lower.7ex\hbox{E}\kern-.125emX}}

\begin{document}

\title{
\vspace*{0.15in}  
Following the Clues: Experiments on Person Re-ID using Cross-Modal Intelligence
\thanks{\textsuperscript{1}Robert Aufschläger, Michael Heigl, Fabian Bally and Martin Schramm are with Deggendorf Institute of Technology, Deggendorf, Germany}
\thanks{\textsuperscript{2}Youssef Shoeb and Azarm Nowzad are with Continental AG, Berlin, Germany}
}

\author {
    \IEEEauthorblockN{
        Robert Aufschläger\textsuperscript{1,}\IEEEauthorrefmark{1}\orcidlink{0009-0004-0986-3504}\thanks{\IEEEauthorrefmark{1}Robert Aufschläger is the corresponding author, his email address is robert.aufschlaeger@th-deg.de}, 
        Youssef Shoeb\textsuperscript{2}\orcidlink{0009-0007-0606-387X}, 
        Azarm Nowzad\textsuperscript{2}, 
        Michael Heigl\textsuperscript{1}\orcidlink{0000-0001-7303-113X}, 
        Fabian Bally\textsuperscript{1}\orcidlink{0009-0004-5656-5008}, \\
        Martin Schramm\textsuperscript{1}\orcidlink{0000-0001-6206-2969}
    }
}

\maketitle

\begin{abstract}
The collection and release of street-level recordings as Open Data play a vital role in advancing autonomous driving systems and AI research. However, these datasets pose significant privacy risks, particularly for pedestrians, due to the presence of Personally Identifiable Information (PII) that extends beyond biometric traits such as faces. In this paper, we present \emph{cRID}, a novel cross-modal framework combining Large Vision-Language Models, Graph Attention Networks, and representation learning to detect textual describable \uline{c}lues of PII and enhance person \uline{r}e-\uline{id}entification (Re-ID). Our approach focuses on identifying and leveraging interpretable features, enabling the detection of semantically meaningful PII beyond low-level appearance cues. We conduct a systematic evaluation of PII presence in person image datasets. Our experiments show improved performance in practical cross-dataset Re-ID scenarios, notably from Market-1501 to CUHK03-np (detected), highlighting the framework’s practical utility.
Code is available at \url{https://github.com/RAufschlaeger/cRID}.
\end{abstract}



\section{Introduction}

The proliferation of Open Data has significantly accelerated progress in artificial intelligence by providing diverse and representative datasets. However, the increasing openness of such data, combined with advancements in camera resolution, raises significant privacy concerns, particularly with visual recordings on persons in public spaces such as streets or parking areas. 
Datasets may expose Personally Identifiable Information (PII), introducing ethical and legal risks related to person Re-Identification (Re-ID). This process involves identifying the same individual across multiple non-interfering cameras and locations by using features like biometrics, pose, clothing, appearance, and body shape to match their identity in different frames.

According to Recital 26 of the European General Data Protection Regulation~(GDPR), the feasibility of identifying individuals, taking into account the time, cost, and technological capabilities involved, must be considered when assessing the need for data protection~\cite{gdpr}. Consequently, a risk analysis of the potential of Re-ID in open datasets is necessary.

\begin{figure}
    \centering
    \includegraphics[width=\columnwidth]{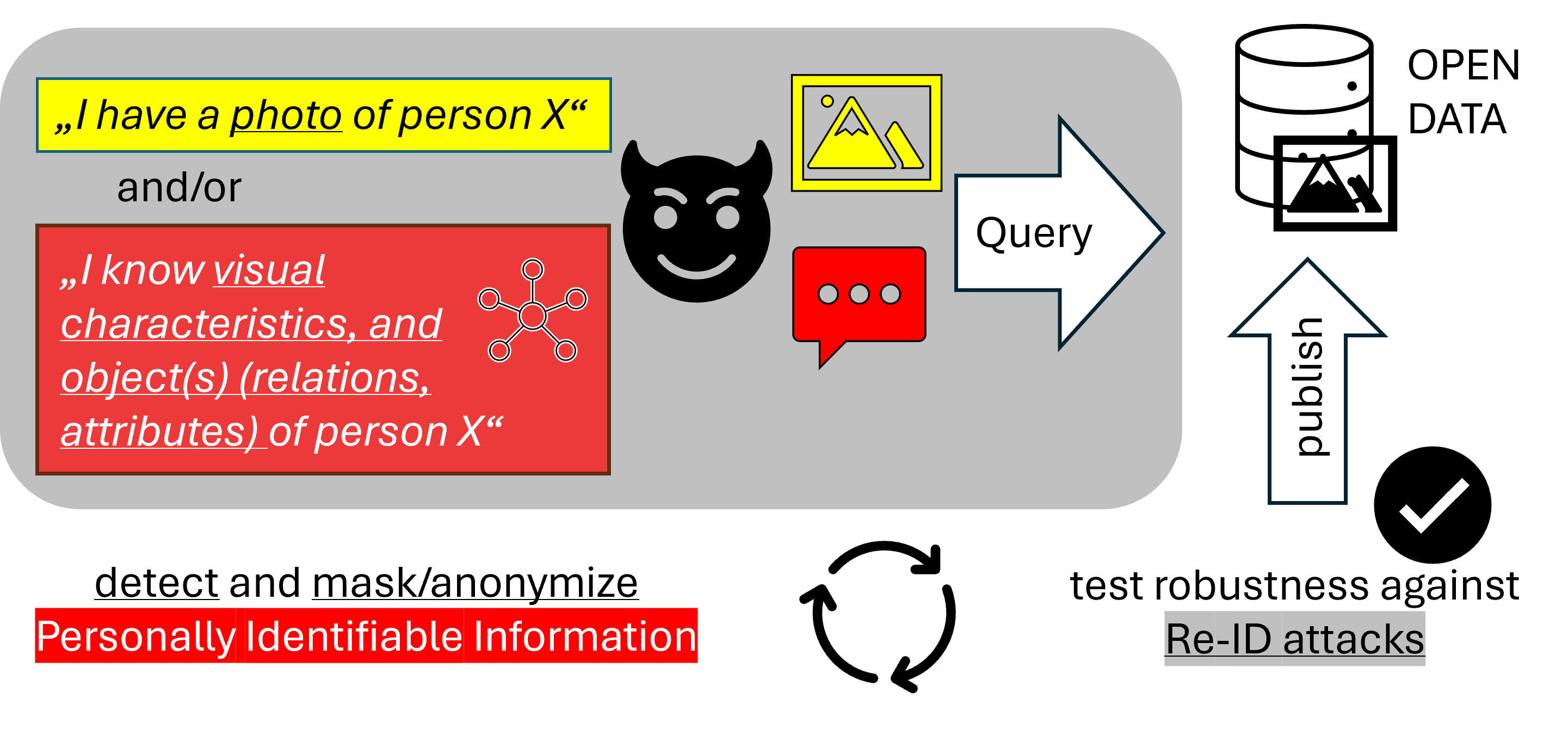}
    \caption{Illustration of a vision-language threat scenario in the context of publishing as Open Data. The gray box represents potential Re-ID attack vectors, where an adversary queries using photographs or graph-structured textual descriptions. The bottom and right elements indicate the essential anonymization and protection processes required for compliant data release. While vision- or language-retrievals can enable attacks, (potentially) interpretable vision and language processing methods can also aid in detecting and anonymizing sensitive data.}
    \label{fig:threat}
\end{figure}

With recent advances in contrastive learning capabilities via image-language pretraining, retrieval methods allow malicious actors to exploit both visual and textual information to retrieve data and re-identify individuals in published image and video datasets~\cite{CLIP-ReID, CILP-FGDI, 10304579}, making Re-ID attacks increasingly feasible. Therefore, it is essential to detect and anonymize PII to mitigate these privacy risks. At the same time, such powerful dual-modality models can be repurposed to enhance privacy protection. By combining language understanding with visual recognition, these systems can improve the semantic interpretation of PII and enable systematic evaluation of Re-ID risks and anonymization needs (Figure \ref{fig:threat}).

On one hand, a potential attacker might use both -- vision and language -- information to retrieve image data. However, conversely, this same synergy between vision and language -- particularly when implemented in Large Vision-Language Models (LVLMs) -- can also be leveraged for privacy protection, such as detecting PII semantics or assessing Re-ID risks through cross-modal alignments.

Building on recent progress in Natural Language Processing and Computer Vision, LVLMs have been successfully applied to a wide range of tasks, including visual question answering and captioning~\cite{deitke2024molmopixmoopenweights}. Based on Large Language Models (LLMs) and vision encoders, these models integrate textual information with visual data, enabling more context-aware understanding based on both modalities. 

LVLMs are foundation models that are pre-trained on huge datasets spanning diverse domains. Their ability to generalize across tasks and contexts is particularly valuable for cross-domain challenges commonly encountered in Re-ID~\cite{CILP-FGDI}.  

In this work, we explore how LVLMs, when combined with graph-based representations, can aid in identifying high-level, interpretable features relevant to PII. Specifically, we examine how LVLM-generated textual scene graphs can enhance the interpretability and risk analysis of Re-ID scenarios. We pose the following \emph{research question
(RQ)}: \emph{\textbf{Can LVLMs be used to construct compact, semantically meaningful scene graph representations of PII in person bounding boxes, and how effective are these representations for assessing Re-ID risk?}}

To address this, we introduce \textbf{cRID} (Clues-guided Re-ID), a cross-modal framework that jointly utilizes image encoders and LVLM-generated scene graphs within a Graph Attention Network (GAT). Our contributions are as follows:

\begin{itemize}
    \item We propose leveraging textual scene graphs from LVLMs to create compact, natural-language descriptions of visual PII.

    \item We present \textbf{cRID}, a novel Re-ID framework that fuses visual embeddings with LVLM-based textual graph structures via a GAT for enhanced identity representation learning.

    \item We extend the strong Re-ID baseline \cite{bagoftricks} with our compact (directed) graph representation and use the GAtt method \cite{gatt} to derive interpretable edge attribution for visual-textual features, modeling realistic threat scenarios for privacy analysis.
    
\end{itemize}

\section{Related Works}

Recent advancements in deep learning have improved Re-ID performance, with state-of-the-art methods increasingly focusing on cross-modal representation and distance metric learning \cite{CLIP-ReID, CILP-FGDI, app14188279, 10204874} while handling prevalent challenges such as occlusions \cite{8486568, 10054607, sym15040906, Yao2024AAGNetAG} or cloth-changing, c.f. \cite{10203842}.

\subsection{VLM-backed Person Re-ID}

VLMs have emerged as a promising approach for Re-ID by leveraging synergies between visual and textual modalities.
Recent works in this domain can broadly be categorized into \emph{image-based} and \emph{text-based} approaches.

\paragraph{Image-based}
Li \textit{et al.} \cite{CLIP-ReID} introduce CLIP-ReID, an approach that leverages the contrastive language-image pre-trained model CLIP \cite{radford2021learning} for Re-ID tasks, specifically addressing the common scenario where concrete text labels are absent. The key idea is to exploit CLIP's cross-modal description ability by learning a set of ID-specific text tokens that provide ambiguous descriptions for each identity. 
While this enables attention-based interpretability, it offers limited semantic grounding and struggles with fine-grained discrimination and domain shifts.

To address these challenges, CILP-FGDI~\cite{CILP-FGDI} proposes a three-stage framework for domain generalization. It includes: 
1. Fine-tuning of CLIP's image encoder to learn Re-ID-specific features 
2. Domain-invariant prompt learning via gradient reversal and learnable domain tokens
3. Bidirectional guidance through contrastive alignment and triplet loss.
This setup improves cross-domain robustness, though prompts remain abstract and not semantically interpretable.
More recently, LVLM-ReID~\cite{wang2024largevisionlanguagemodelsmeet} extends this line by employing LVLMs to extract semantic pedestrian tokens guided by textual instructions.
While offering richer semantics than CLIP-based methods, the approach lacks explicit disentanglement of specific visual attributes critical for Re-ID, instead aggregating them into a single token.

\paragraph{Text-based}
Text-based person Re-ID seeks to retrieve a person’s image given a free-form text. 
Leveraging CLIP’s cross-modal capabilities, CFine~\cite{10304579} introduces a modular framework: (1) multi-level global feature learning to extract discriminative local details within each modality, (2) cross-grained feature refinement to establish coarse-level cross-modal correspondences, and (3) fine-grained correspondence discovery to capture detailed visual-textual relationships. CFine offers interpretability via attention maps, image-word, and sentence-patch similarity scores, but is limited in modeling complex interdependencies between attributes.
Similarly, IRRA \cite{10204874} addresses modality heterogeneity and intra-identity variation incorporating CLIP. 
Unlike approaches with explicit part alignment, IRRA models local relations implicitly. While it excels in word-level semantic matching, it struggles with phrase-level understanding, where contextual relations across multiple words are crucial.

\subsection{Graph-backed Person Re-ID}

Recent Re-ID research addresses occlusions by using Graph Neural Networks to reason about visible and occluded body parts, sometimes incorporating textual attributes or descriptions, and attention mechanisms.

For image-based occluded person Re-ID, Huang \textit{et al.} \cite{10054607} propose the Reasoning and Tuning GAT to focus on reliable features. It learns a representation that jointly reasons visible body parts and compensates occluded parts. 
Their approach offers interpretability through activation heatmaps and introduces visibility scores to interpret visible and occluded part features. However, it does not involve textual explanations and focuses solely on visibility.
AAGNet \cite{Yao2024AAGNetAG} tackles occlusions by mapping manually labeled attributes (e.g., hat color, bag style) to word vectors and fusing them with body part features via graph convolution. While effective in crowded scenarios, it relies on predefined body parts and manual annotations, limiting scalability. 

For text-based person Re-ID, BAMG~\cite{BAMG25} proposes a Bottleneck Attention and Masked Graph Modeling framework combining CLIP-based image and text encoders with a bottleneck transformer for multimodal fusion.
Similarly, A-GANet \cite{Liu2019DeepAG} leverages adversarial learning to align visual and textual scene graphs, modeling object attributes and relations through graph attention convolution layers. This facilitates cross-modal feature alignment in a shared latent space. However, the method relies on pre-defined textual queries and does not interpret attention weights.

\subsection{Summary of limitations in current approaches}

Current approaches in Re-ID, especially those leveraging (L)VLMs and graph-based methods, have made significant progress in addressing occlusion and cross-modal alignment. However, they still face several key limitations. Many methods lack explicit, interpretable mappings between learned features and semantic attributes, instead relying on abstract prompts (vague descriptions) or aggregated tokens (combined data points), which provide limited semantic grounding and hinder interpretability.
Moreover, current models often struggle with fine-grained attribute discrimination, making it difficult to distinguish between similar attributes. Additionally, these models frequently fail in capturing complex relational or phrase-level semantics, which are critical for robust scene understanding and privacy-aware Re-ID.

\section{Methodology}
\label{sec:methodology}

\paragraph{Task definition}

Our problem statement corresponds to \emph{supervised Re-ID}. 

In supervised person Re-ID with representation learning, the objective is to learn a feature embedding model that maps each image to a $d$-dimensional space. Embeddings of images belonging to the same identity should be close, while those of different identities should be distant. The Re-ID process then consists of: Computing embeddings, comparing query and gallery embeddings, ranking gallery images based on similarity scores.

When training Re-ID models, the given labeled datasets are commonly used as follows.

Let
$
\mathcal{D} = \{(x_i, y_i)\}_{i=1}^N
$
represent a dataset, where $x_i \in \mathbb{R}^{H \times W \times C}$ is a bounding box of a person images and $y_i \in \{1, 2, ..., P\}$ is the person ID (label) with $P$ equals the total number of unique identities.
The dataset is split into three subsets: 1) \emph{Training set}: $
   \mathcal{D}_{\text{train}} = \{(x_i, y_i)\}_{i=1}^{N_{\text{train}}}
   $, where identities in this set are disjoint from those in the test set
2) \emph{Query set}: $
   \mathcal{D}_{\text{query}} = \{(x_j, y_j)\}_{j=1}^{N_{\text{query}}}
   $
   containing images for which matching needs to be performed 3) \emph{Gallery set}: $
   \mathcal{D}_{\text{gallery}} = \{(x_k, y_k)\}_{k=1}^{N_{\text{gallery}}}
   $
   containing candidate images to match against. The gallery set includes at least one image for each identity present in the query set. 
The query and gallery sets correspond to the test set.

\paragraph{System Architecture}

\begin{figure}[ht]
    \centering
    \includegraphics[width=0.49\textwidth]{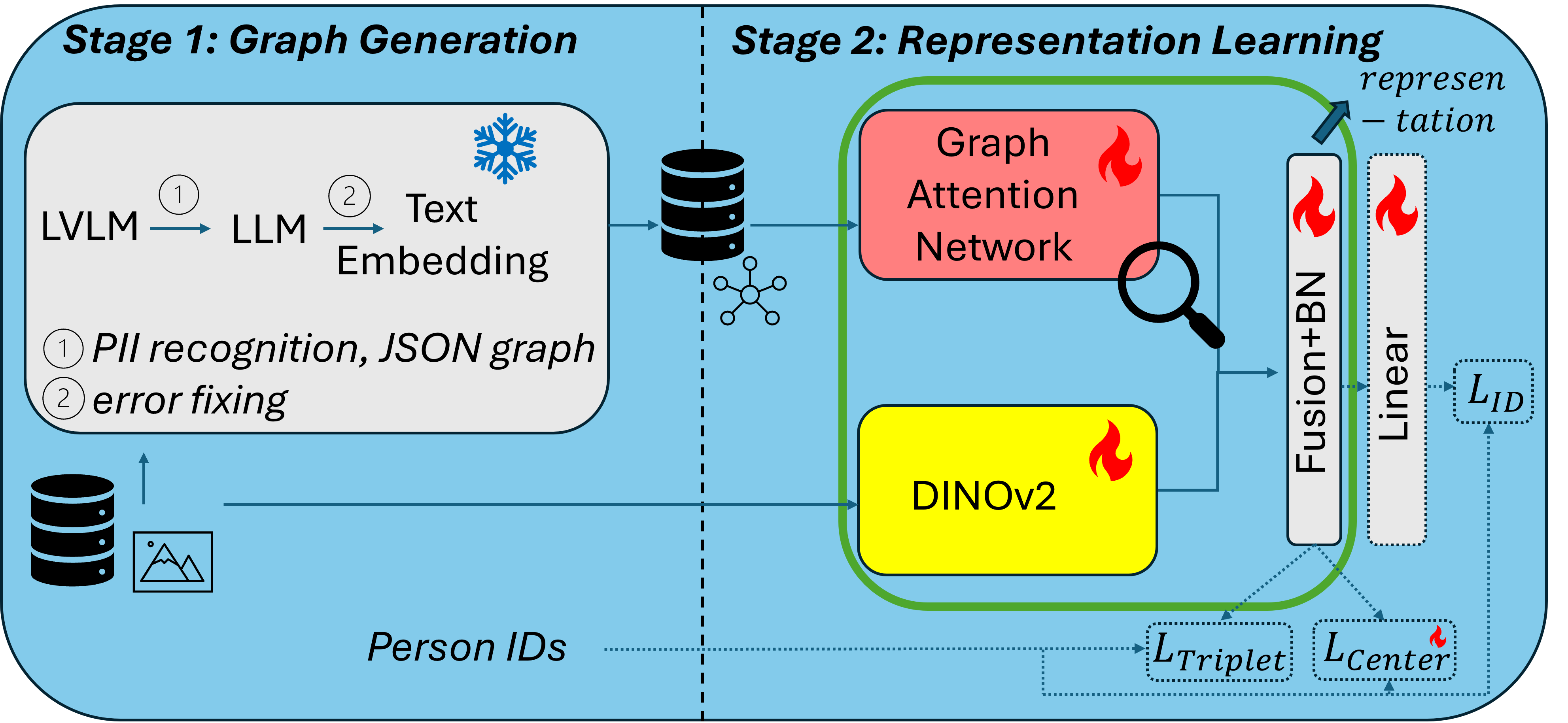}
    \caption{Overview of the proposed two-stage Re-ID framework (\textbf{cRID}). The components are explained in detail in Section \ref{sec:methodology}. While GAT or DINOv2 can be used individually as the encoding backbone, the green border indicates their use as a joint backbone.}
    \label{fig:architecture}
\end{figure}

We implement a two-stage approach with LVLM-based graph generation (Stage 1) and a two-branch model architecture for representation learning (Stage 2) (Figure \ref{fig:architecture}). In Stage 2, in one branch the LVLM-extracted semantic scene graphs are processed in a trainable two-layered GAT, and in the other branch visual features are extracted using a pre-trained DINOv2~\cite{oquab2023dinov2} or ResNet50~\cite{ResNet50} backbone. The two outputs are merged into an aligned representation by a trainable fusion network and concatenated with a classification layer.

\paragraph{DINOv2 Visual Encoding}

To extract robust visual features, we employ META's DINOv2~\cite{oquab2023dinov2}. We experiment with ViT-S, ViT-B, and ViT-L variants (patch size $= 14$). As a baseline, we use features extracted from ResNet50~\cite{ResNet50}.

\paragraph{LVLM}

To capture high-level semantics, we use Molmo-7B\footnote{\url{https://huggingface.co/allenai/Molmo-7B-O-0924}}~\cite{deitke2024molmopixmoopenweights}, an LVLM optimized for visual question answering and captioning. It is built on OLMo-7B-1024 and OpenAI’s CLIP ViT-L/14 encoder. We apply it to generate a \emph{scene graph} $G=(O,E)$ for each image, using the prompt shown in Figure \ref{fig:graph-generation-prompt}.

\begin{figure}[ht]
    \centering
    \begin{tcolorbox}[colback=hellgrauHintergrundZwei, colframe=dunkelblaugruen60heller, title=, sharp corners=south, fonttitle=\bfseries, fontupper=\small]
    \textit{Describe the detailed visual characteristics of the person in the photo that could be used to re-identify the person. Create a scene graph. Use the JSON format like in the example shown below and only output the JSON:}
    \begin{lstlisting}[basicstyle=\ttfamily\scriptsize]
    {
      "nodes": [
        { "id": "person", "attributes": 
            ["...", "...", "..."] },
        { "id": "...", "attributes": 
            ["...", "...", "..."] }
      ],
      "edges": [
        { "source": "...", "target": "...", 
            "relation": "..." },
        { "source": "...", "target": "...", 
            "relation": "..." }
      ]
    }
    \end{lstlisting}
    \textit{Ensure nodes, attributes, and edges are well-structured. Ensure that the JSON is valid, and do not output additional information. In the output, use only English language. Nodes consist of an id and an attributes list. Edges consist of a source, a target, and a relation. Use only up to 1000 tokens.}
\end{tcolorbox}
    \caption{Graph generation prompt.}
    \label{fig:graph-generation-prompt}
\end{figure}

\paragraph{LLM}

\begin{figure}[ht]
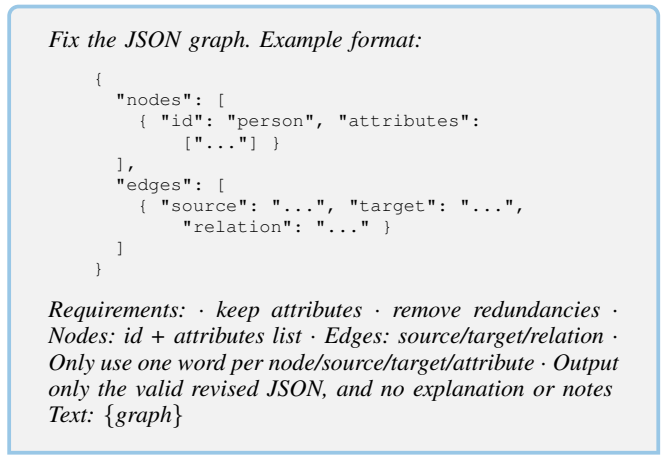

    \centering
    \begin{tcolorbox}[colback=hellgrauHintergrundZwei, colframe=dunkelblaugruen60heller, title=, sharp corners=south, fonttitle=\bfseries, fontupper=\small]
    \textit{Fix the JSON graph.}
    \textit{Example format:}
    \begin{lstlisting}[basicstyle=\ttfamily\scriptsize]
    {
      "nodes": [
        { "id": "person", "attributes": 
            ["..."] }
      ],
      "edges": [
        { "source": "...", "target": "...", 
            "relation": "..." }
      ]
    }
    \end{lstlisting}
    \textit{Requirements:}
    \textit{$\cdot$ keep attributes}
    \textit{$\cdot$ remove redundancies}
    \textit{$\cdot$ Nodes: id + attributes list}
    \textit{$\cdot$ Edges: source/target/relation}
    \textit{$\cdot$ Only use one word per node/source/target/attribute}
    \textit{$\cdot$ Output only the valid revised JSON, and no explanation or notes}
    
    \textit{Text: \{graph\}}
    \end{tcolorbox}
    \caption{Fix JSON graph prompt.}
    \label{fig:fix-graph-prompt}
\end{figure}

To robustly recover valid scene graphs from malformed JSON input, the system combines a Large Language Model (LLM) (phi4 (14b)~\cite{abdin2024phi-})-based repair function (\texttt{fix\_graph}) with a rule-based post-processing step (\texttt{fix\_malformed\_json}). We opted for an LLM in the fixing step due to its superior adaptability and ability to handle a broader, more complex range of malformed outputs often produced by LVLMs. While simpler rule-based systems are efficient for known patterns, they often struggle with the nuanced and inconsistent errors that can arise from advanced model outputs. The LLM's contextual understanding enables intelligent corrections that go beyond rigid syntactic rules, effectively repairing issues such as invalid edge sets or misformatted attributes that may not strictly violate JSON syntax but still disrupt the intended structure (prompt in Figure \ref{fig:fix-graph-prompt}). After the LLM proposes a fix, \texttt{fix\_malformed\_json} applies systematic syntax cleanups -- removing extraneous markers, correcting improper escaping, and restructuring edge definitions to ensure the result is both semantically coherent and syntactically valid. Notably, the fixing was rarely necessary in practice: only $30$ out of $36,036$ graphs in Market-1501, $3$ out of $14,097$ in CUHK03-NP (detected), and $6$ out of $14,096$ in CUHK03-NP (labeled) required repair, indicating high initial annotation quality and the robustness of the overall processing pipeline.


\paragraph{Text Embedding}

To get numerical graph representations, the textual described edges and nodes are converted into a $386$ dimensional numerical representation. Thereby, the all-MiniLM-L6-v2 embedding model~\cite{wang-etal-2021-minilmv2}  was chosen for its optimized balance of computational efficiency and semantic encoding capability in text processing. This $6$-layer distilled model ($22$M parameters, $384$-dim embeddings) employs multi-head self-attention relation distillation. The approach is particularly effective for short text semantic matching.

\paragraph{Graph Attention Network}

Subsequently, after obtaining graph data from the LVLM, where attributes are added as additional nodes, our approach employs a GAT, which is a form of neural network architecture that operates on graph data, to dynamically weight node relationships in representation learning, prioritizing critical connections while suppressing noise. 
GATs are adept at handling missing or noisy edges, and, specifically, we use two cascaded GATv2 convolutional layers~\cite{brody2022how} ($1$ attention head, with self loops, incorporating edge attributes) using PyTorch Geometric's (PyG's) \texttt{GATv2Conv} module\footnote{\url{https://pytorch-geometric.readthedocs.io/en/latest/generated/torch_geometric.nn.conv.GATv2Conv.html}, 25/02/2025}, with attention mechanisms applied between connected node pairs $(u,v)$ in the graph. GATv2 can enhance performance if directionality is significant \cite{fi16090318}. We reverse the information flow to \texttt{target$\rightarrow$source} to establish a person node-centric information flow.
Notably, flipping the edge directions in experiments resulted in significantly worse performance.

The \texttt{GATv2Conv} layers are followed by a global max pooling layer and a layer normalization module to output representations (dim: $128$). The attention weights $\alpha_{uv}$ are computed at each layer as

\begin{equation}
\alpha_{uv} =
\operatorname{softmax}_v\Bigl(
  \mathbf{a}^\top
  \operatorname{LReLU}\Bigl(
    \mathbf{W}_s \mathbf{x}_u
    + \mathbf{W}_t \mathbf{x}_v
    + \mathbf{W}_e \mathbf{e}_{uv}
  \Bigr)
\Bigr),
\end{equation}

\noindent where
\begin{itemize}
\item $\mathbf{W}_s, \mathbf{W}_t$: Shared learnable weight matrix for source/target node features
\item $\mathbf{W}_\mathbf{e}$: Shared learnable weight matrix for edge features
\item $x_u, x_v$: Feature vectors of nodes $u, v \in O$ in the graph (dim: $384$)
\item $\mathbf{e}_{uv}$: Feature vector of edge $(u \rightarrow v) \in E$ (dim $384$)
\item $\mathbf{a}$: Attention vector parameterizing node interactions
\item $\operatorname{LReLU}(x)$: Leaky Rectified Linear Unit activation
\end{itemize}

Then, each node’s representation is updated via attention-weighted aggregation of its neighbors.
GATv2Conv addresses standard GAT's static attention limitation \cite{brody2022how} while enhancing gradient flow through LeakyReLU activations and normalization. In our setup, we intentionally configure \texttt{GATv2Conv} with single attention heads to enable direct visualization and interpretation of attention patterns.

\paragraph{Fusion}

In the Multi Layer Perceptron (MLP) based fusion both image and graph representations are integrated. The MLP consists of a single linear layer to process the data to produce compact $128$-dimensional representations.

\paragraph{Loss}

We optimized a combined loss function based on the strong baseline findings \cite{bagoftricks}, where the loss is calculated as a combination of losses
\begin{equation}
    \mathcal{L} = \mathcal{L}_{Triplet} + \lambda \mathcal{L}_{Center} + \mathcal{L}_{ID}.
    \label{eq:loss}
\end{equation}

The triplet loss $\mathcal{L}_{Triplet}$ \cite{TripletLoss} enforces discriminative learning of representations, ensuring that instances of the same class are mapped closer in feature space while increasing the separation between different classes. 

The center loss $\mathcal{L}_{Center}$~\cite{CenterLoss} maintains learnable class centers in feature space and penalizes distance between features and their class centers. The $\lambda$ parameter balances with triplet loss~\cite{bagoftricks}. 

Additionally, ID Loss $\mathcal{L}_{ID}$ is used, as it better utilizes the training data and makes the model more robust \cite{classificationloss}.
Our person classification is based on predicting the person identity labels on the training set. The module takes the representation vectors from the fusion module, normalizes them with \texttt{BatchNorm} (BN), and then uses a single \texttt{Linear} layer to predict the person's identity. The linear layer is only used in training, where its output is used to calculate the \texttt{CrossEntropyLoss}.

\section{Experiments}

\subsection{Setup}

\paragraph{Hardware}

All experiments were conducted on a virtual machine equipped with $12$ cores of an AMD EPYC 9534 64-Core Processor, $64$GB of RAM, and $2$ NVIDIA L40S GPUs ($2\times48$GB VRAM), running CUDA 12.4 and NVIDIA driver version 550.144.03. Proxmox was used as the Hypervisor.

\paragraph{Configurations}
\label{par:configs}

Our findings are built upon the codebase from \cite{bagoftricks}. We use the same setup for each of the models and datasets and do not apply any kind of data augmentation. Specifically, we do not apply flipping and random erasing due to negative impact on cross-dataset evaluation. We apply re-ranking \cite{cuhk03-np}, use the features after batch normalization and normalize returned features.
Input images are resized to $252 \times 126$ pixels to maintain compatibility with ViT's $14\times14$ patch structure (except when using ResNet50 backbone). In the applied ResNet50 backbone, the last spatial downsampling stride in the final residual block is set to $1$, again following \cite{bagoftricks}. Training employs Adam optimization (warm-up learning rate with base learning rate: $LR=0.00035$) over $120$ epochs, using a batch size of $64$ images/graphs (sample $4$ instances per identity). The loss function (Eq. \eqref{eq:loss}) uses $\lambda=0.0005$, and is enhanced by label smoothing.

\paragraph{Datasets}
We use the person bounding box datasets Market1501 and CUHK03-np (detected).

\textbf{Market1501}~\cite{market1501} contains $32,668$ annotated images of $1,501$ identities, captured from $6$ cameras in a real-world surveillance setting in front of a supermarket in daylight, incl. high number of distinct identities, plenty of well-separated clothing-based visual clues, clear poses and consistent backgrounds. The person bounding boxes are automatically detected and cropped by the Deformable Part Model (DPM).
The query/ folder contains $3,368$ images and represents $750$ identities. The bounding\_box\_train directory holds $12,936$ images. This set includes $751$ identities, with multiple images captured from different camera angles per individual. The bounding\_box\_test directory consists of $19,732$ images, forming the gallery set for retrieval incl. $750$ identities, along with an additional background class to introduce realistic challenges in identifying individuals correctly.

\textbf{CUHK03-np (detected)}~\cite{cuhk03-np, cuhk03} contains $14,096$ images of $1,467$ identities, each captured from two cameras on the CUHK campus -- averaging $4.8$ images per camera. The new protocol divides the dataset into a training set ($767$ identities) and a testing set ($700$ identities). The precise split comprises: $1,400$ query images, $5,328$ gallery (test) images, and $7,368$ training images for the labeled set, and $1,400/5,332/7,365$ for the detected split. We use the noisy (detected) version, where bounding boxes are generated using the DPM pedestrian detector.

\paragraph{Evaluation Metrics}
To assess the performance of our person Re-ID framework, we report Rank-1 accuracy (R@1), Rank-5 accuracy (R@5), and mean Average Precision (mAP). 
While the mAP evaluates the overall ranking quality by considering the precision of all correct matches in the list, the R@k metrics directly measure the retrieval performance at the top ranks, which is more critical in real-world applications. These metrics measure the likelihood that at least one correct match appears within the top-$k$ retrieved items.

\subsection{Results}

Table \ref{tab:reid_results} presents the evaluation results of various backbone models and their hybrid versions with GAT across Market1501, CUHK03-np (detected), and cross-dataset scenarios.

\begin{table*}[ht]
    \centering
    \caption{Person Re-ID performance evaluation. We compare pretrained baseline models DINOv2 and ResNet50 with the GAT, and hybrid variants on Market1501, and CUHK03-np (detected) datasets. In addition, cross-dataset evaluations are reported. 
    }
    \label{tab:reid_results}
    \resizebox{\textwidth}{!}{%
    \begin{tabular}{l|ccc|ccc|ccc|ccc}
        \hline
        \multirow{2}{*}{\textbf{Backbone}}
         & \multicolumn{3}{c|}{\textbf{Market1501}} 
         & \multicolumn{3}{c|}{\textbf{CUHK03-np (detected)}} 
         & \multicolumn{3}{c|}{\textbf{M$\rightarrow$C}}
         & \multicolumn{3}{c}{\textbf{C$\rightarrow$M}} \\
        \cline{2-13}
         & \textbf{R@1} & \textbf{R@5} & \textbf{mAP} 
         & \textbf{R@1} & \textbf{R@5} & \textbf{mAP} 
         & \textbf{R@1} & \textbf{R@5} & \textbf{mAP} 
         & \textbf{R@1} & \textbf{R@5} & \textbf{mAP} \\
        \hline

        \texttt{ResNet50} & \textbf{94.3} & \textbf{96.7} & \textbf{91.3} & 61.6 & 73.1 & 65.6 & 9.9 & 15.4 & 11.8 & 54.8 & 65.0 & 35.9 \\
        \texttt{DINOv2\_vits14} & 90.6 & 94.7 & 86.1 & 66.4 & \textbf{77.7} & 69.6 & 5.6 & 10.5 & 7.0 & 49.3 & 60.9 & 32.1 \\
        \texttt{DINOv2\_vitb14} & 93.0 & 96.6 & 89.1 & \textbf{67.1} & 77.6 & \textbf{70.0} & 7.1 & 12.8 & 8.4 & \textbf{57.5} & 68.0 & 39.6 \\
        \texttt{DINOv2\_vitl14} & 92.5 & 96.2 & 89.2 & 63.5 & 74.9 & 66.8 & 7.1 & 12.9 & 8.8 & 57.3 & \textbf{68.9} & \textbf{40.2} \\

                \hline

        \texttt{GAT} & 31.8 & 53.6 & 19.7 & 11.5 & 23.3 & 11.5 & \textbf{12.4} & \textbf{23.0} & \textbf{12.2} & 25.9 & 46.3 & 14.4 \\

                \hline

        \texttt{GAT} / \texttt{ResNet50}             & 47.5 & 68.5 & 41.2 & 19.9 & 36.3 & 21.3 & 6.9 & 14.4 & 7.7 & 5.7 & 13.3 & 3.2 \\
        \texttt{GAT} / \texttt{DINOv2\_vits14}             & 82.2 & 90.7 & 75.7 & 48.9 & 64.7 & 49.7 & 8.4 & 15.7 & 9.0 & 13.2 & 25.1 & 8.3 \\
        \texttt{GAT} / \texttt{DINOv2\_vitb14}             & 82.7 & 90.5 & 76.8 & 56.9 & 71.8 & 57.4 & 10.7 & 17.8 & 11.7 & 25.8 & 42.8 & 16.3 \\
        \texttt{GAT} / \texttt{DINOv2\_vitl14} & 82.3 & 91.0 & 75.8 & 55.1 & 71.1 & 57.2 & 10.0 & 17.3 & 11.1 & 31.8 & 50.0 & 19.6 \\

        \hline
    \end{tabular}%
    }
\end{table*}

\paragraph{Quantitative Analysis}

The ResNet50 baseline performs strongly on the Market1501 dataset, achieving a high Rank-1 accuracy of 94.3\% and mAP of 91.3\%. However, its performance degrades on the noisier CUHK03-np (detected) dataset (61.6\% R@1, 65.6\% mAP), and drops significantly in cross-dataset evaluations (e.g., Market1501 $\rightarrow$ CUHK03-np (detected) (M$\rightarrow$C), R@1 9.9\%). DINOv2-based vision transformers generally outperform ResNet50 on CUHK03-np. Particularly, DINOv2\_vitb14 achieves 67.1\% R@1 on CUHK03-np, compared to 61.6\% for ResNet50. Cross-dataset generalization also improves slightly: DINOv2\_vitb14 achieves 57.5\% R@1 when transferring from CUHK03-np (detected) to Market1501 (C$\rightarrow$M). Incorporating GAT can improve performance in cross-dataset evaluation. GAT/DINOv2\_vitb14 improves M$\rightarrow$C performance to 10.7\% R@1 and C$\rightarrow$M to 25.8\% R@1, outperforming pure backbone baselines. 
This supports the hypothesis that graph-based modeling of semantic scene graphs introduces domain-invariant, interpretable identity cues.
Transitioning from DINOv2\_vits14 (small) to vitb14 (base) improves overall performance, while moving to DINOv2\_vitl14 (large) yields only marginal gains, suggesting diminishing returns when scaling up model size without specific domain adaptation. The GAT model trained independently of visual features performs poorly across all datasets (e.g., Market1501 R@1 31.8\%), underlining that visual feature extraction remains critical for person Re-ID.

\paragraph{Influence of dataset characteristics}

The performance differences across datasets can be attributed to their distinct properties:

\textbf{Market1501}:  
Captured under controlled outdoor conditions with six fixed surveillance cameras, Market1501 provides clean detections, consistent backgrounds, clear poses, and rich clothing-specific clues. The automatic DPM detector introduces some noise but mostly produces consistent crops. With a high number of distinct identities ($751$ train / $750$ test) and ample intra-class variation, the dataset favors visual extractors like DINOv2 and ResNet50 that rely on global appearance features.
Market1501 images exhibit minimal occlusions and accurate detections, enabling the LVLM to extract well-structured, meaningful graphs. These graphs allow the GAT to model fine-grained semantic relations (e.g., ``red backpack'', ``blue jacket'') that correlate strongly with identity.

\textbf{CUHK03-np (detected)}:  
In contrast, CUHK03-np is captured in a university campus setting with higher viewpoint and pose variability. Detections are noisier, with frequent blur, occlusion, and background clutter. The dataset features fewer distinct identities ($767$) and more challenging conditions. Although DINOv2 ViT-S still outperform ResNet50, the noisy detections lead to weaker graphs. Consequently, GAT-enhanced models do not show as large a margin of improvement on CUHK03 compared to Market1501.

\paragraph{Cross-Dataset Generalization}

Cross-dataset evaluation reveals the impact of domain shifts:  
Models trained on Market1501 suffer performance drops when tested on CUHK03 due to differences in pose, background complexity, and camera viewpoints.
Among the DINOv2 backbones, \text{ViT-B} and \text{ViT-L} perform similarly and achieve better results than \text{ViT-S} on CUHK03, with all DINOv2 variants outperforming ResNet50. This demonstrates the advantage of larger vision transformers when handling more complex and noisy datasets. Interestingly, even the smaller ViT-S model achieves better cross-dataset generalization compared to ResNet50, highlighting the strength of self-supervised pretraining. However, CUHK03 also has more occlusion and intra-class variation, which challenges self-supervised methods not fine-tuned on that domain and still attention-based cannot fully bridge the domain gap. Here, GAT-based hybrids help by modeling high-level semantic attributes that remain stable across domains (such as color, carried objects, or accessories).

The cross-dataset evaluation from Market1501 to CUHK03 (M$\rightarrow$C) is particularly challenging and realistic, as it reflects the difficulty of transferring models from clean to noisy domains. In contrast, when training on CUHK03 and testing on Market1501 (C$\rightarrow$M), the hybrid GAT-based models perform worse. This is mainly because CUHK03 provides fewer training samples, noisier detections, and weaker graphs, limiting the benefits of graph-based conclusions during transfer to a cleaner target domain. As a result, strong pretrained visual backbones, such as DINOv2 ViT-B and ViT-L, perform better in this setting, since robust visual feature extraction becomes the dominant factor when graph-based cues are unreliable.

\paragraph{Interpretability}

\begin{figure*}[ht]
\centering
\includegraphics[width=1\linewidth]{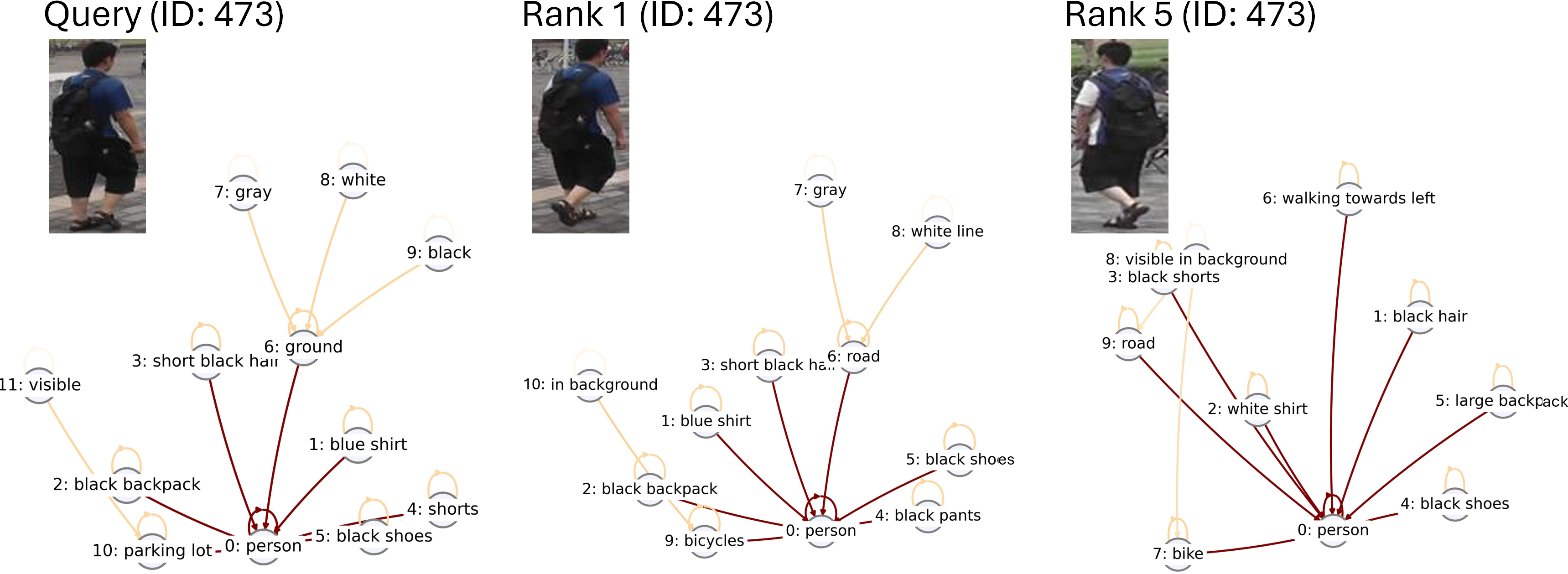}
\caption{Example Re-ID query with interpretability analysis using attribution scores from GAtt \cite{gatt} on node '0: person'. This figure illustrates a query image and its top retrieved matches,
alongside the textual graph representations (nodes and attributes) and their corresponding
attribution scores. Due to GAtt's methodology, non-contributing nodes are omitted in these graph plots. The model weights were loaded from \texttt{GAT / DINOv2\_vitb14} -- trained on
Market1501/bounding\_box\_train \cite{market1501}.
The query image (0473\_c4s2\_050698\_00.jpg) is from the query split, and the retrievals are from the gallery set bounding\_box\_test. Edge attributes are omitted for better readability. The nodes are numerated based on the order obtained from the LVLM.}
\label{fig:enter-label}
\end{figure*}

Figure \ref{fig:enter-label} exemplifies the ability to extract and visualize semantically meaningful, graph-structured attributes of PII from visual data. These attributes, listed in order as obtained from the LVLM, are then transformed into explicit textual graphs (e.g., ``black backpack'', blue shirt'', ``short black hair''), providing an interpretable understanding of what can make a person re-identifiable beyond pose and face biometrics.
Finding these graphs can enhance the safety of public datasets from a privacy perspective in several ways:
\begin{itemize}
\item \textbf{Complementary Risk Assesment}: The explicit textual attributes within these graphs allow for a systematic and granular evaluation of Re-ID risks. This goes beyond merely detecting faces to understanding specific visual cues (like clothing details or carried objects) that an adversary could exploit. By understanding these specific semantic links, the framework can model a realistic Re-ID threat scenario that consider both visual and textual attack vectors. This capability helps data publishers assess how individuals might be identified and evaluate datasets for their robustness against sophisticated Re-ID attempts, which is crucial for compliant data release.
\item \textbf{Targeted Anonymization}: Once these interpretable PII attributes are identified through the graph representations, data publishers can implement more precise and effective anonymization strategies. Instead of broad blurring or pixelation that might destroy valuable data utility, specific features contributing to re-identifiability can be targeted for masking or alteration. This allows for a balance between data utility for research and individual privacy protection.
\end{itemize}
To further invastigate this interpretability, we test edge attribution using GAtt~\cite{gatt}, a hyperparameter-free and deterministic explanation method specifically designed for GATs. Experiments (Figure \ref{fig:enter-label}, bottom) indicate an averaging effect in attribution scores (similar color = similar score). Subsequently, in our setting GAtt does not provide insights about most important nodes, however filters person-related information due to edge directions. In future work, a possible solution might be to penalize close attention weights in the loss function and/or favor simpler losses.

\section{Conclusions and discussion}

This work explores the potential of LVLMs to generate compact, interpretable scene graph representations of PII within person image bounding boxes, and assesses their capacity to provide insights into realistic Re-ID threats.
Our proposed framework \textbf{cRID} demonstrates that LVLMs can effectively extract semantically meaningful, graph-structured attributes from visual data, bridging a critical gap in current research, where previous approaches often lacked including automatically generated explicit and interpretable personal attributes. Here, the combination of LVLM-generated semantic graphs with GAT provides a novel approach to Re-ID that aligns with realistic threat scenarios.

The second part of our research question concerned insights into realistic Re-ID threats. Our cross-dataset evaluations revealed that semantically-grounded representations can indeed provide practical insights about Re-ID risks. Through extensive experiments on real-world Re-ID benchmarks, we found that integrating LVLM-generated scene graphs with visual features in a graph-based architecture (via GAT) provides practical insights about Re-ID risks by modeling identity through interpretable textual descriptions. Notably, this approach performed especially well on the challenging Market1501 to CUHK03-np (detected) cross-domain generalization scenario ($10.7$\% R@1 compared to $7.1$\% for DINOv2\_vitb14 alone), underscoring its practicality for Open Data privacy analysis. The framework thus advances the state of the art by enabling privacy-sensitive Re-ID analysis, modeling both visual and textual threat vectors, and facilitating systematic risk assessment of PII exposure.

Nonetheless, our study identified several limitations, including computational inefficiency for real-time use (due to the LVLM's GPU requirements and inference time), lack of end-to-end optimization, and a primary focus on local rather than global contextual cues. The latter might hinder generalization to more challenging Re-ID scenarios such as cloth-change or long-term Re-ID. Despite these challenges, our research demonstrates that the synergy of language and vision models can bring new levels of transparency, and privacy-awareness to Re-ID which is used to test anonymizations (Figure \ref{fig:threat}).

\section{Future work}

We plan to continue working on these results. Enhancing computational efficiency and enabling real-time applications requires model distillation and better integration of spatial-temporal context, especially for video data. Also, we aim to explore bias mitigation techniques, such as causal reasoning for cloth-change scenarios~\cite{10203842}, to focus the LVLM-based graphs on truly identity-relevant features. Applying our method to a broader variety of datasets -- including occluded person and vehicle Re-ID, and scenarios with distractor data -- will further test its robustness.

On the privacy front, we envision developing methods to rate and quantify Re-ID risks based on LVLM-generated PII attributes, potentially supporting both pre- and post-anonymization analysis. Furthermore, recognized PII could also enhance (negative) prompt-based anonymization techniques. Finally, as LVLMs and visual backbones continue to improve, we see opportunities to boost performance further through different pre-training datasets~\cite{pass2024}, better models, prompt tuning, sophisticated data augmentation, and hyperparameter optimization.

\section*{Ethical Statement}
While adversarial prompts are employed in the LVLMs to probe Re-ID risks, their sole purpose is to stress-test and improve anonymization methods. By demonstrating how open-source frameworks can trivialize privacy-invasive applications, this work underscores the dual-use implications of AI and advocates for proactive safeguards -- including privacy-by-design principles -- to mitigate harm while advancing anonymization research.
This research uses only publicly available datasets.

\section*{Acknowledgement}
This work results from the just better DATA project supported by the German Federal Ministry for Economic Affairs and Climate Action (BMWK), grant number 19A23003H.

\bibliographystyle{IEEEtran}
{\small
\bibliography{library}}

\begin{thebibliography}{10}
\providecommand{\url}[1]{#1}
\csname url@samestyle\endcsname
\providecommand{\newblock}{\relax}
\providecommand{\bibinfo}[2]{#2}
\providecommand{\BIBentrySTDinterwordspacing}{\spaceskip=0pt\relax}
\providecommand{\BIBentryALTinterwordstretchfactor}{4}
\providecommand{\BIBentryALTinterwordspacing}{\spaceskip=\fontdimen2\font plus
\BIBentryALTinterwordstretchfactor\fontdimen3\font minus \fontdimen4\font\relax}
\providecommand{\BIBforeignlanguage}[2]{{%
\expandafter\ifx\csname l@#1\endcsname\relax
\typeout{** WARNING: IEEEtran.bst: No hyphenation pattern has been}%
\typeout{** loaded for the language `#1'. Using the pattern for}%
\typeout{** the default language instead.}%
\else
\language=\csname l@#1\endcsname
\fi
#2}}
\providecommand{\BIBdecl}{\relax}
\BIBdecl

\bibitem{gdpr}
``Regulation (eu) 2016/679 of the european parliament and of the council of 27 april 2016 on the protection of natural persons with regard to the processing of personal data and on the free movement of such data (general data protection regulation),'' \url{https://eur-lex.europa.eu/eli/reg/2016/679/oj}, 2016, official Journal of the European Union, L 119, 4 May 2016.

\bibitem{CLIP-ReID}
\BIBentryALTinterwordspacing
S.~Li, L.~Sun, and Q.~Li, ``Clip-reid: exploiting vision-language model for image re-identification without concrete text labels,'' in \emph{Proceedings of the Thirty-Seventh AAAI Conference on Artificial Intelligence and Thirty-Fifth Conference on Innovative Applications of Artificial Intelligence and Thirteenth Symposium on Educational Advances in Artificial Intelligence}, ser. AAAI'23/IAAI'23/EAAI'23.\hskip 1em plus 0.5em minus 0.4em\relax AAAI Press, 2023. [Online]. Available: \url{https://doi.org/10.1609/aaai.v37i1.25225}
\BIBentrySTDinterwordspacing

\bibitem{CILP-FGDI}
H.~Zhao, L.~Qi, and X.~Geng, ``Cilp-fgdi: Exploiting vision-language model for generalizable person re-identification,'' \emph{IEEE Transactions on Information Forensics and Security}, pp. 1--1, 2025.

\bibitem{10304579}
S.~Yan, N.~Dong, L.~Zhang, and J.~Tang, ``Clip-driven fine-grained text-image person re-identification,'' \emph{IEEE Transactions on Image Processing}, vol.~32, pp. 6032--6046, 2023.

\bibitem{deitke2024molmopixmoopenweights}
\BIBentryALTinterwordspacing
M.~Deitke \emph{et~al.}, ``Molmo and pixmo: Open weights and open data for state-of-the-art vision-language models,'' 2024. [Online]. Available: \url{https://arxiv.org/abs/2409.17146}
\BIBentrySTDinterwordspacing

\bibitem{bagoftricks}
H.~Luo, Y.~Gu, X.~Liao, S.~Lai, and W.~Jiang, ``Bag of tricks and a strong baseline for deep person re-identification,'' in \emph{Proceedings of the IEEE/CVF Conference on Computer Vision and Pattern Recognition (CVPR) Workshops}, June 2019.

\bibitem{gatt}
\BIBentryALTinterwordspacing
Y.-M. Shin, S.~Li, X.~Cao, and W.-Y. Shin, ``Faithful and accurate self-attention attribution for message passing neural networks via the computation tree viewpoint,'' \emph{Proceedings of the AAAI Conference on Artificial Intelligence}, vol.~39, no.~19, pp. 20\,461--20\,469, Apr. 2025. [Online]. Available: \url{https://ojs.aaai.org/index.php/AAAI/article/view/34254}
\BIBentrySTDinterwordspacing

\bibitem{app14188279}
M.~Ma, J.~Wang, and B.~Zhao, ``A multi-scale graph attention-based transformer for occluded person re-identification,'' \emph{Applied Sciences}, vol.~14, no.~18, 2024.

\bibitem{10204874}
D.~Jiang and M.~Ye, ``Cross-modal implicit relation reasoning and aligning for text-to-image person retrieval,'' in \emph{2023 IEEE/CVF Conference on Computer Vision and Pattern Recognition (CVPR)}, 2023, pp. 2787--2797.

\bibitem{8486568}
J.~Zhuo, Z.~Chen, J.~Lai, and G.~Wang, ``Occluded person re-identification,'' in \emph{2018 IEEE International Conference on Multimedia and Expo (ICME)}, 2018, pp. 1--6.

\bibitem{10054607}
M.~Huang, C.~Hou, Q.~Yang, and Z.~Wang, ``Reasoning and tuning: Graph attention network for occluded person re-identification,'' \emph{IEEE Transactions on Image Processing}, vol.~32, pp. 1568--1582, 2023.

\bibitem{sym15040906}
M.~Zhu and H.~Zhou, ``Ecreid: Enhancing correlations from skeleton for occluded person re-identification,'' \emph{Symmetry}, vol.~15, no.~4, 2023.

\bibitem{Yao2024AAGNetAG}
\BIBentryALTinterwordspacing
S.~Yao, K.~Pan, T.~Wang, Z.~Zheng, J.~Jin, and C.~Hu, ``Aagnet: Attribute-aware graph-based network for occluded pedestrian re-identification,'' \emph{IEEE Transactions on Consumer Electronics}, vol.~70, pp. 6580--6588, 2024. [Online]. Available: \url{https://api.semanticscholar.org/CorpusID:272388031}
\BIBentrySTDinterwordspacing

\bibitem{10203842}
Z.~Yang, M.~Lin, X.~Zhong, Y.~Wu, and Z.~Wang, ``Good is bad: Causality inspired cloth-debiasing for cloth-changing person re-identification,'' in \emph{2023 IEEE/CVF Conference on Computer Vision and Pattern Recognition (CVPR)}, 2023, pp. 1472--1481.

\bibitem{radford2021learning}
A.~Radford \emph{et~al.}, ``Learning transferable visual models from natural language supervision,'' in \emph{International conference on machine learning}.\hskip 1em plus 0.5em minus 0.4em\relax PMLR, 2021, pp. 8748--8763.

\bibitem{wang2024largevisionlanguagemodelsmeet}
\BIBentryALTinterwordspacing
Q.~Wang, B.~Li, and X.~Xue, ``When large vision-language models meet person re-identification,'' 2024. [Online]. Available: \url{https://arxiv.org/abs/2411.18111}
\BIBentrySTDinterwordspacing

\bibitem{BAMG25}
K.~Cheng, W.~Zou, H.~Gu, and A.~Ouyang, ``Bamg: Text-based person re-identification via bottlenecks attention and masked graph modeling,'' in \emph{Computer Vision -- ACCV 2024}, M.~Cho, I.~Laptev, D.~Tran, A.~Yao, and H.~Zha, Eds.\hskip 1em plus 0.5em minus 0.4em\relax Singapore: Springer Nature Singapore, 2025, pp. 384--401.

\bibitem{Liu2019DeepAG}
\BIBentryALTinterwordspacing
J.~Liu, Z.~Zha, R.~Hong, M.~Wang, and Y.~Zhang, ``Deep adversarial graph attention convolution network for text-based person search,'' \emph{Proceedings of the 27th ACM International Conference on Multimedia}, 2019. [Online]. Available: \url{https://api.semanticscholar.org/CorpusID:204837191}
\BIBentrySTDinterwordspacing

\bibitem{oquab2023dinov2}
M.~Oquab \emph{et~al.}, ``Dinov2: Learning robust visual features without supervision,'' \emph{Transactions on Machine Learning Research (TMLR)}, 2023.

\bibitem{ResNet50}
K.~He, X.~Zhang, S.~Ren, and J.~Sun, ``Deep residual learning for image recognition,'' in \emph{Proceedings of the IEEE Conference on Computer Vision and Pattern Recognition (CVPR)}, June 2016.

\bibitem{abdin2024phi-}
\BIBentryALTinterwordspacing
M.~I. Abdin \emph{et~al.}, ``Phi-4 technical report,'' Microsoft, Tech. Rep. MSR-TR-2024-57, 12 2024. [Online]. Available: \url{https://www.microsoft.com/en-us/research/publication/phi-4-technical-report/}
\BIBentrySTDinterwordspacing

\bibitem{wang-etal-2021-minilmv2}
\BIBentryALTinterwordspacing
W.~Wang, H.~Bao, S.~Huang, L.~Dong, and F.~Wei, ``{M}ini{LM}v2: Multi-head self-attention relation distillation for compressing pretrained transformers,'' in \emph{Findings of the Association for Computational Linguistics: ACL-IJCNLP 2021}, C.~Zong, F.~Xia, W.~Li, and R.~Navigli, Eds.\hskip 1em plus 0.5em minus 0.4em\relax Online: Association for Computational Linguistics, Aug. 2021, pp. 2140--2151. [Online]. Available: \url{https://aclanthology.org/2021.findings-acl.188/}
\BIBentrySTDinterwordspacing

\bibitem{brody2022how}
\BIBentryALTinterwordspacing
S.~Brody, U.~Alon, and E.~Yahav, ``How attentive are graph attention networks?'' in \emph{International Conference on Learning Representations}, 2022. [Online]. Available: \url{https://openreview.net/forum?id=F72ximsx7C1}
\BIBentrySTDinterwordspacing

\bibitem{fi16090318}
A.~G. Vrahatis, K.~Lazaros, and S.~Kotsiantis, ``Graph attention networks: A comprehensive review of methods and applications,'' \emph{Future Internet}, vol.~16, no.~9, 2024.

\bibitem{TripletLoss}
F.~Schroff, D.~Kalenichenko, and J.~Philbin, ``Facenet: A unified embedding for face recognition and clustering,'' in \emph{2015 IEEE Conference on Computer Vision and Pattern Recognition (CVPR)}, 2015, pp. 815--823.

\bibitem{CenterLoss}
Y.~Wen, K.~Zhang, Z.~Li, and Y.~Qiao, ``A discriminative feature learning approach for deep face recognition,'' in \emph{Computer Vision -- ECCV 2016}, B.~Leibe, J.~Matas, N.~Sebe, and M.~Welling, Eds.\hskip 1em plus 0.5em minus 0.4em\relax Cham: Springer International Publishing, 2016, pp. 499--515.

\bibitem{classificationloss}
Y.~Zhai, X.~Guo, Y.~Lu, and H.~Li, ``In defense of the classification loss for person re-identification,'' in \emph{Proceedings of the IEEE/CVF Conference on Computer Vision and Pattern Recognition (CVPR) Workshops}, June 2019.

\bibitem{cuhk03-np}
Z.~Zhong, L.~Zheng, D.~Cao, and S.~Li, ``Re-ranking person re-identification with k-reciprocal encoding,'' in \emph{2017 IEEE Conference on Computer Vision and Pattern Recognition (CVPR)}, 2017, pp. 3652--3661.

\bibitem{market1501}
L.~Zheng, L.~Shen, L.~Tian, S.~Wang, J.~Wang, and Q.~Tian, ``Scalable person re-identification: A benchmark,'' in \emph{Proceedings of the 2015 IEEE International Conference on Computer Vision (ICCV)}, ser. ICCV '15.\hskip 1em plus 0.5em minus 0.4em\relax USA: IEEE Computer Society, 2015, p. 1116–1124.

\bibitem{cuhk03}
W.~Li, R.~Zhao, T.~Xiao, and X.~Wang, ``Deepreid: Deep filter pairing neural network for person re-identification,'' in \emph{2014 IEEE Conference on Computer Vision and Pattern Recognition}, 2014, pp. 152--159.

\bibitem{pass2024}
K.~Zhu, H.~Guo, T.~Yan, Y.~Zhu, J.~Wang, and M.~Tang, ``Pass: Part-aware self-supervised pre-training for person re-identification,'' in \emph{Computer Vision -- ECCV 2022}, S.~Avidan, G.~Brostow, M.~Ciss{\'e}, G.~M. Farinella, and T.~Hassner, Eds.\hskip 1em plus 0.5em minus 0.4em\relax Cham: Springer Nature Switzerland, 2022, pp. 198--214.

\end{thebibliography}



\end{document}